\documentclass{article}

\usepackage{arxiv}

\usepackage[utf8]{inputenc} 
\usepackage[T1]{fontenc}    
\usepackage{hyperref}       
\usepackage{url}            
\usepackage{booktabs}       
\usepackage{amsfonts}       
\usepackage{nicefrac}       
\usepackage{microtype}      
\usepackage{lipsum}		
\usepackage{graphicx}
\usepackage{natbib}
\usepackage{doi}

\title{Dull, Dirty, Dangerous: Understanding the Past, Present, and Future of a Key Motivation for Robotics}

\author{ \href{https://orcid.org/0000-0002-6618-0695}{\includegraphics[scale=0.06]{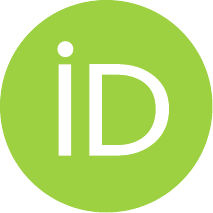}\hspace{1mm}Nozomi Nakajima}\thanks{Authors contributed equally to the paper.} \\
	Robotics and AI Institute\\
	Cambridge, MA 02142 \\
	\texttt{nnakajima@rai-inst.com} \\
	\And
	\href{https://orcid.org/0000-0001-5690-1850}{\includegraphics[scale=0.06]{orcid.pdf}\hspace{1mm}Pedro Reynolds-Cuéllar}$^*$ \\
	Robotics and AI Institute\\
	Cambridge, MA 02142 \\
	\texttt{pcuellar@rai-inst.com} \\
	\And
	\href{https://orcid.org/0009-0006-6912-0547}{\includegraphics[scale=0.06]{orcid.pdf}\hspace{1mm}Caitrin Lynch} \\
	Robotics and AI Institute\\
    Cambridge, MA 02142 \\
    Olin College of Engineering \\
    Needham, MA 02492 \\
	\texttt{clynch@rai-inst.com} \\
	\And
	\href{https://orcid.org/0000-0002-2236-3078}{\includegraphics[scale=0.06]{orcid.pdf}\hspace{1mm}Kate Darling} \\
	Robotics and AI Institute\\
	Cambridge, MA 02142 \\
	\texttt{kdarling@rai-inst.com} \\
}

\renewcommand{\shorttitle}{Dull, Dirty, and Dangerous}

\hypersetup{
pdftitle={Dull, Dirty, Dangerous: Understanding the Past, Present, and Future of a Key Motivation for Robotics},
pdfsubject={cs.CY, cs.RO},
pdfauthor={Nozomi Nakajima, Pedro Reynolds-Cuéllar, Caitrin Lynch, Kate Darling},
pdfkeywords={Dull, Dirty, Dangerous; Labor; Robotics; Automation; Future of Work; Literature Review},
}

\begin{document}
\maketitle

\begin{abstract}
	In robotics, the concept of “dull, dirty, and dangerous” (DDD) work has been used to motivate where robots might be useful. In this paper, we conduct an empirical analysis of robotics publications between 1980 and 2024 that mention DDD, and find that only 2.7\% of publications define DDD and 8.7\% of publications provide concrete examples of tasks or jobs that are DDD. We then review the social science literature on “dull,” “dirty,” and “dangerous” work to provide definitions and guidance on how to conceptualize DDD for robotics. Finally, we propose a framework that helps the robotics community consider the job context for our technology, encouraging a more informed perspective on how robotics may impact human labor. 
\end{abstract}

\keywords{Dull, Dirty, Dangerous \and Labor \and Robotics \and Automation \and Future of Work \and Literature Review}

\section{Introduction}
There is frequent public concern that robots will replace humans in the workplace, leading to worries about rising unemployment and decreased access to meaningful work. Within the robotics community, the phrase “dull, dirty, and dangerous” (DDD) has become conventional justification to motivate the types of tasks or jobs for which robots might be useful---by doing work that's undesirable for people to do. A classic example of a DDD job is one of “repetitive physical labor on a steaming hot factory floor involving heavy machinery that threatens life and limb”~\citep{takayama_beyond_2008}. However, while identifying DDD work may seem intuitive, there are complex (and sometimes hidden) social, economic, and cultural factors that can determine the value of different kinds of labor. What exactly constitutes “dull” work, and who makes that inference? How do we define “dirty” work in ways that acknowledge both physical and social dimensions of work? On what basis do we classify work as “dangerous”? 

This paper argues that exploring DDD in more depth by drawing on social science literature about work helps us better understand the impact of robotic technology on human labor. We propose a new DDD framework to help the robotics community examine which types of tasks or jobs robots are well suited to perform and desirable to automate. While DDD is only one of many possible approaches to evaluating the desirability of automation, its popularity in robotics makes it a natural starting point. Our framework provides an actionable tool for robotics to guide technological choices in more socially beneficial ways.

First, we empirically review nearly 1,000 papers published in the field of robotics between 1980 and 2024. Our analysis reveals that, while DDD is widely used to motivate research, very few papers define these terms or provide provenance citations. 

Second, we contextualize this empirical data by reviewing social sciences literature on “dull,” “dirty,” and “dangerous” work. We find that this literature identifies a nuanced relationship between work and the surrounding physical and social environments. Notably, the social sciences highlight the importance of worker perspective for accurately establishing whether work is considered dull, dirty, and/or dangerous.  

Finally, we propose a new framework that provides clear definitions and guidance for examining jobs through the DDD lens. This framework allows us to expand on the conventional narrative of DDD in robotics. It also helps robotics go beyond our primary focus of advancing technical capabilities, and to consider the job context that our technology will impact. 

In summary, this paper makes three primary contributions to the field of HRI. 1. An empirical analysis of the use of the DDD concept across robotics literature; 2. A review of the definitions and concepts of these terms in social sciences, as well as what we believe to be key considerations for robotics; 3. A new framework encouraging the critical exploration of tasks and jobs in robotics. These contributions aim to support the robotics community in considering the downstream effects of automation, by  broadening how we imagine robots supporting human labor.

\section{Related Work}

Our study builds on literature in robotics and social sciences that investigates the impact of robotics on workers. While some scholars and industry analysts argue that robots may provide an important solution to revitalizing the economy with manufacturing~\citep{leigh_robots_2020, manjoo_how_2017, muller_americas_2025, oppenheimer_robots_2019}, others, concerned about potential negative outcomes, focus on the impact on human relationships and the meanings of human labor when people work alongside or are displaced by robots~\citep{amodei_machines_2024, holzer_understanding_2022, nikolova_robots_2024}. Our research aims to bridge this divide by bringing an interdisciplinary lens to the question of where and how robotic technology might be most beneficial to human labor and society.  

Social scientists have studied the impacts of robotics in the wider context of automation and emerging technologies. Regarding the question of robots taking people’s jobs,~\citep{acemoglu_robots_2017} examine U.S. data from 1990 to 2007 and show that introducing robots may lead to reduction in employment-to-population ratio, particularly for workers without college degrees. In a study of the changing nature of work and the future of work,~\citep{walley_robots_2020} considers what the introduction of robots and AI means symbolically for people in the U.S. concerned about job loss. Broader global perspectives, like the edited collection by~\citep{paus_confronting_2018}, consider the economic and political implications of AI, digital connectivity, and robotics. While a collection by Pink et al. on “everyday automation”~\citep{pink_everyday_2022} is not focused on robotics per se, it provides insights into how workers respond to emerging automation-related technologies and how workplace technology impacts workers. Within this collection,~\citep{berg_hate_2022} argues that ideas about what is suitable for automation are embedded in our values about the present and future of human activity and work. As Berg notes, “a technology or system for process automation ... is also a sociotechnical prism through which the past, present and future of work are imagined and re-constructed.” Our proposed framework also examines robotics through a sociotechnical prism, moving beyond a purely technical focus.

In HRI studies on how robots and humans work together, scholars have considered worker perspectives when deploying robots. Scholars have identified positive and negative work attributes that can guide the integration of robots in the workplace; analyzed how the introduction of robots has led to new social interactions among workers and altered workplace organizational structures; and noted the interaction of the social capabilities of robots and the social expectations of humans working with them~\citep{welfare_consider_2020,cheon_robots_2022,sauppe_social_2015}. 

Our study builds both on HRI and social science approaches to robotics and workers, foregrounding the sociotechnical context for technology deployment and identifying an opportunity to bolster the suitability and impact of robots. Our project is most closely connected to other HRI studies that examine the concept of DDD~\citep{riek_future_2025, takayama_beyond_2008}. 

~\citep{riek_future_2025} debunk five myths about supposed worker-positive reasons for automation: the first myth is that nobody wants DDD jobs anyway. We share these authors' interest in questioning assumptions and broadening what voices and experiences are considered in determining what to automate. Riek and Irani refer to scholars who have highlighted the different “sociotechnical imaginaries” between end users and engineers, such as~\citep{eriksson_politicising_2024},~\citep{rahm_educational_2019}, and~\citep{toll_sociotechnical_2022}. We join these researchers in calling for a more deliberate consideration of the social implications of robotics. Our paper builds upon these previous works by providing empirical evidence on the extent to which robotics researchers use the concept of DDD. 

Our work also seeks to provide value by offering a new DDD framework for analyzing this key robotics concept.~\citep{takayama_beyond_2008} argue that the question of what occupations robots should be permitted to do is “\textit{subjective}, depending on the norms of the community” [italics in original]. Aligned with this view, our proposed framework explicitly calls for the inclusion of multiple perspectives. While we agree with Takayama et al.'s position that the DDD heuristic is just one of many ways to frame a conversation about the future of work and robotics, we suggest that the lens of DDD is worth retaining as an entry point, in particular due to its widespread use in robotics. 

The DDD framework we propose in the paper specifically draws attention to the worker perspective. While hourly wages, benefits, and hours are important dimensions of a job, literature shows that workers also highly value non-wage amenities~\citep{mas2017, maestas2023}. Workplace dignity (which includes autonomy on the job, coworker relationships, and quality of managerial supervision) is particularly important to workers~\citep{hodson2001, reich2018}. Robots in the workplace can promote or undermine workplace dignity depending on how the technology is implemented~\citep{gibson2023}, and our new DDD framework provides a concrete approach to consider this worker perspective.

\section{Review of the use of DDD in Robotics Papers}

\subsection{Methodology}
We began by developing a dataset of peer-reviewed research publications that explicitly mentioned “dull,” “dirty,” and “dangerous” (DDD) work.\footnote{While other D words sometimes accompany this triad (e.g., “demeaning,” “degrading,” “demanding,” “difficult,” “dear,” among others), our data collection efforts revealed that the vast majority of papers that used these alternative terms also included at least one of the three core DDD terms.} Our inclusion criteria for this dataset were peer-reviewed papers that: (1) were published in engineering/robotics conferences and journals; (2) mentioned robots and/or robotics; (3) were written in English. Because we were interested in discovering how the academic literature refers to and employs DDD, we made no differentiation in terms of how and where these terms appeared within the text. Our data collection process was informed by prior systematic review work within the HRI/HCI literature~\citep{cooper_systematic_2022,taylor_cruising_2024}. The process of developing our dataset was as follows:  

\subsubsection{Data Collection: Information Sources and Search Strategies} 
Our search focused on two publication databases of interest, the ACM Digital Library and the IEEE Xplore digital library, in addition to Dimensions. (Dimensions is an aggregation platform that indexes multiple sources including CrossRef, PubMed, Directory of Open Access Journals, Open Citation Data, and multiple preprint repositories [e.g., arXiv].) We ran multiple search iterations to refine our search terms across  platforms. We continuously vetted random samples from each query to ensure that they met our inclusion criteria. Iterations helped balance sensitivity and compliance with inclusion criteria. For example, we benchmarked results between searches that required the presence of all three D terms and key terms of interest (e.g., dull, dirty, dangerous, robots, robotics) against searches that specified word proximity (e.g., “dull, dirty, dangerous” AND “robot” within three words of each other). Whenever the database allowed, we specified searches across specific catalogs (e.g., “40 Engineering” or “46 Information and Computing Sciences”) to narrow down the results. We did not include grey literature in our search strategy. While we performed a full search of news publications using Media Cloud~\citep{roberts_media_2021}, an aggregation platform, we found relevant results sparse and difficult to narrow down in relation to specific academic domains. Detailed queries and overall search strategy are in an adapted PRISMA flowchart in the Appendix. 

\subsubsection{Data Collection: Screening and Data Extraction}
We imported all search results into Zotero in order to retain and organize metadata for each entry. We then prepared the corpus for further screening, removing duplicates and texts that were not peer-reviewed (e.g., transcripts of keynotes, reports). Due to copyright access, we excluded books and book chapters. We also removed any entries not in the English language. After screening, 919 entries remained for data extraction and analysis.

To prepare for analysis, we programmatically extracted all sentences containing our key terms, along with the three preceding and following sentences to ensure sufficient context. The second author reviewed all extracted data for accuracy and further cleaned the dataset, ensuring the reliability of the process and that all excerpts met our eligibility criteria. Next, we extracted qualitative data from the dataset annotating each excerpt for: (1) jobs/occupations referenced; (2) specific tasks mentioned (e.g., environmental monitoring, search and rescue); (3) definitions of DDD work; and (4) academic field of publication. These annotations were performed manually by the second author and, since they did not require interpretation of the data or lead to the creation of a codebook for qualitative analysis, we did not require an inter-coder reliability metric. 

\subsubsection{Data Analysis}
Following methods from content analysis~\citep{krippendorff_content_nodate}, we quantified our findings, reporting on the number of papers that provided citations for DDD. We also report on the number of times authors provided definitions for any of the DDD terms and extracted these definitions. Using annotations on the dataset, we report on the number of jobs mentioned, as well as popular jobs used by authors as exemplary of DDD concepts. Here, we differentiated jobs (e.g., pilot, farmer) from tasks (e.g., surveillance, plant monitoring). We separate these results by academic field.

\subsection{Results}
\subsubsection{Prevalence of the DDD terms}
\begin{figure}
    \includegraphics[width=\columnwidth, alt={Figure is a stacked area chart showing the cumulative count of articles from 1980 to 2024 across three categories: All sources (919 total), Papers mentioning jobs (80 total), and Papers with definitions (25 total). The vertical axis shows Cumulative count of articles, ranging from 0 to 1,000. The horizontal axis shows Year from 1980 to 2024. The chart demonstrates minimal activity from 1980 to approximately 2004, with cumulative counts remaining near zero. Growth begins around 2004, accelerates significantly after 2012, and increases steeply from 2016 onward, reaching approximately 920 articles by 2024. Papers with definitions comprise 2.7 percent of all sources, shown in the bottom layer. Papers mentioning jobs comprise 8.7 percent of all sources, shown in the middle layer. The remaining articles, comprising the majority, are shown in the top layer. The overall pattern shows exponential growth in literature on this topic, particularly in the past decade.}]{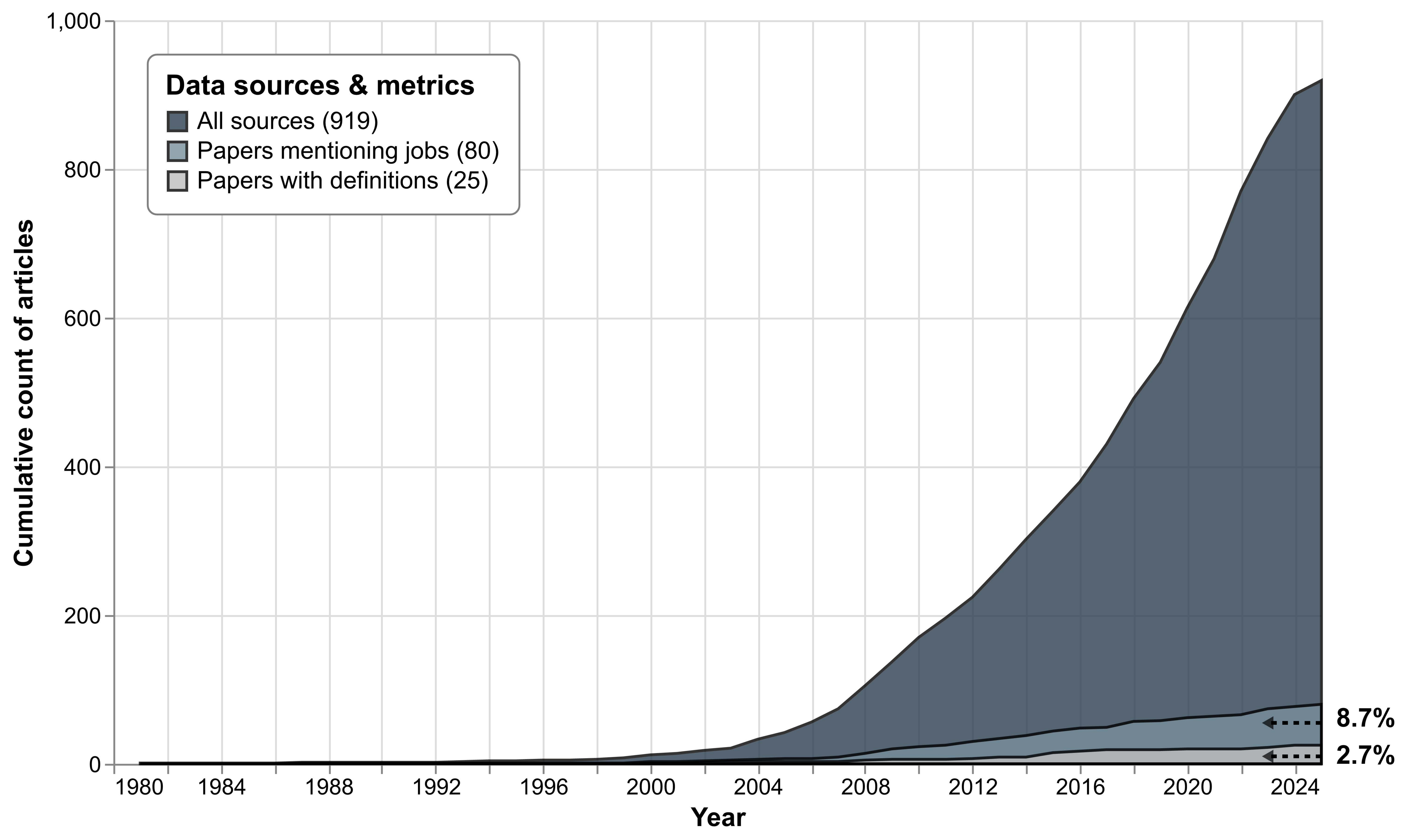}
    \caption{Graph aggregating yearly counts of mentions of the DDD concept in papers across all three databases reviewed}
    \label{fig:ddd-over-time}
\end{figure}

Our data shows that the prevalence of DDD in the literature has steadily increased since 1981 and picked up pace starting in 2008. Figure~\ref{fig:ddd-over-time} shows this progression. Given that DDD is often used to justify or motivate the application of robotics, we examined the types of jobs and tasks associated in our dataset. We found that only 80 publications (8.7\%) mentioned specific jobs when referring to DDD. These authors referenced 96 different types of jobs as exemplary of either dull, dirty, or dangerous jobs with “pilot,” “soldier,” and “farmer” as the most common categories. While other authors offered tasks as representative of DDD, often these examples lacked specificity, making it difficult to parse what aspects of a tasks they deemed as dull, dirty, or dangerous. Examples included mentions of robots performing “industrial manufacturing,” “space exploration,” or “home care” tasks. 

\subsubsection{DDD definitions and contexts of use}
Only 25 publications (2.7\%) provided a formal definition of one or more of the DDD concepts. The definitions we found varied widely. Some authors referred to DDD jobs as “social roles that users may not be interested in taking on themselves”~\citep{luriaSocialBoundariesPersonal2020} or “those not best suited for a human operator due to the circumstances in which they are to be undertaken”~\citep{daveModelingControlNonlinear2015}. A small group of authors offered specific definitions of each term, although in most cases, they were narrowly specific to the military context~\citep{okcuUtilizationUnmannedAircraft2016, cullenMissionSupportDrones2017}. For these and additional examples, see Appendix Table 1.

We also found that only 194 publications (21\%) provided citations with their mentions of DDD. The prominent sub-fields in these citations are UAV/drones (36.1\%), critical and social examinations of robots (16.5\%), and manufacturing (8.2\%). Notably, a sizable portion of the papers citing sources for DDD concepts appeared from 2020 onward (49.5\%). While prior years showed steady growth in citing sources, 2020 represents a notable increase. Upon reviewing these citations, we found that only 15 provided any formal definition of one or more of the DDD concepts. Appendix Table 1 lists papers in our dataset that support or define DDD, along with the definitions provided by the authors. 

\section{Review of Social Science Literature}
\label{sec:socsci}

Our empirical results showed that in robotics research, few publications provided concrete examples or definitions of DDD. We then turned to social sciences literature, where scholars have extensively studied the nature of labor, to see if we could import a more robust definition of DDD into the field of robotics.

\subsection{Methodology}

We conducted a literature review to synthesize how social scientists have studied and defined “dull,” “dirty,” and “dangerous” work. We took an interdisciplinary approach, spanning anthropology, economics, political science, psychology, and sociology, to collect fundamental works and contemporary research. This approach allowed us to capture the multifaceted nature of the terms. Rather than a systematic literature review with predefined search and study selection criteria, we provide a broad synthesis of fundamental social science literature, in order to represent the diverse scholarly landscape concerning the nature of work. 

In the section below, we take each of the terms in DDD and provide: an agreed-upon definition (in \textbf{bold}) from the social sciences, as well as (i) a review of how the term has been studied in the social sciences, and (ii) our key considerations for robotics.

\subsection{Dangerous Work}

\subsubsection{Social science review} “Dangerous work” is defined as \textbf{occupations or tasks that result in injury or risk of harm}. The definition stems from standards in occupational safety and health statistics, which play a key role in identifying and characterizing dangerous work. One of the most common ways to measure dangerous work is using occupational injury, defined as “personal injury resulting from an unexpected and unplanned occurrence arising out of or in connection with work”~\citep{international_labour_organization_quick_2020}. An occupational injury can be fatal (where death occurred within one year of the occupational accident) or non-fatal with lost work time. For ease of interpretation and comparability across jobs and geographies, occupational injuries are typically reported per 10,000 workers in the reference group~\citep{international_labour_organization_implementing_2023}. 

Another approach to measuring dangerous work focuses on exposure to hazardous risk factors using job exposure matrices (JEM). JEM are cross-tabulations of job titles and risk factors, with cells indicating the intensity of exposure to a particular risk factor for each job title~\citep{bouyer_retrospective_1993}. JEM is used not only for physical and chemical hazards---which are often restricted to certain occupational groups---but also for psychosocial or work organization hazards (e.g., job control, job demands, and social support at work) that exist in all occupations~\citep{choi_developing_2020}. 

Statistics on occupational injuries and risk exposure come from a variety of data sources, including administrative records collected by agencies (e.g., insurance records, labor inspection records, death registers) and surveys collected for research (e.g., establishment/firms surveys, household surveys)~\citep{international_labour_organization_quick_2020}. For both types of data sources, understanding how and why data were reported, collected, and verified is critical to assessing the validity of the data and the statistical measures that derive from them.

\subsubsection{Considerations for robotics} 

First, in areas where there is likely to be underreporting, there may be more opportunities for robotics to address dangerous work (see Section~\ref{sec:framework:example} for a waste and recycling industry example). The underreporting of occupational injuries and risk is well documented, with some studies estimating up to 70\% of occupational injuries missing in administrative databases~\citep{leigh_estimate_2004}. Multiple points in the reporting process contribute to this undercount: workers may be afraid to report to their supervisors injuries and risks~\citep{lipscomb_safety_2013}; workers may not be able to afford time off, and only injuries leading to time off are identified as occupational injuries in many reporting systems~\citep{laske_employees_2022, azaroff_occupational_2002}; workers may receive care outside of their workers’ compensation system~\citep{spieler_re_2016, propublica_insult_2015}; and workers may not immediately perceive a connection between their health problems and their job~\citep{azaroff_occupational_2002, international_labour_organization_call_2023}. When using occupational safety and health statistics to understand which work is dangerous, figures need to be interpreted alongside broader labor market and legal contexts, such as the prevalence of labor inspections, informal employment, and social protection coverage~\citep{siqueira_effects_2014, international_labour_organization_quick_2020}. 

Second, those looking to automate dangerous work should consult multiple data sources and long-term trends when considering occupational injuries and risks related to dangerous work. Statisticians generally warn against identifying certain occupations as the “most dangerous” in a given year based on any one measure. Because these rates can be highly unstable, injury rates are not calculated for many occupations that have a relatively small number of work injuries and employment~\citep{bureau_of_labor_statistics_national_2024}. To illustrate this point: very few workers are employed in the occupation of “elephant trainer,” so a small number of fatal injuries would make the fatal injury rate extremely high. However, in most years, this occupation incurs no deaths, making it among the least at risk for fatal injuries~\citep{toscano_dangerous_1997}.    

Third, researchers should be aware that currently, there are few data on occupational injuries and risk factors disaggregated by worker and work characteristics such as gender, migration status, formal/informal employment, and work activities~\citep{international_labour_organization_quick_2020}. This lack of disaggregated data on occupational risks means that work-related hazards may be disproportionately concentrated among certain populations. If we can get a better understanding of how work-related risks are concentrated across society, this could shed light on additional groups of workers that could benefit from improved safety measures in the workplace. For example, occupational safety and health research has historically focused on male-dominated industries such as manufacturing, construction, and transportation. As a result, safety benchmarks for various hazards were established using men as reference, although recent research suggests that these benchmark levels could still be unsafe for women~\citep{perez_invisible_2019}. Jobs in these male-dominated industries also pose specific physical risks such as falls and strikes, whereas traditionally female-dominated industries (e.g., caregiving, healthcare, service) place workers at higher risk for musculoskeletal disorders due to repetitive tasks and long hours of standing~\citep{messing_be_2003, biswas2022differences}. Searching for (and supporting efforts to collect and report) disaggregated data can help reveal how different workers are exposed to occupational risks and identify opportunities to promote and extend safety measures to all workers.

\subsection{Dirty Work}
\subsubsection{Social science review} “Dirty work” is defined as \textbf{occupations or tasks that are physically, socially, or morally tainted}~\citep{hughes_good_1962}.  The social science literature makes clear that dirty work is not only reserved for \textit{physically} dirty jobs, but encompasses \textit{stigmatized} work that is perceived as undesirable or degrading. Work that is physically tainted involves garbage, death, effluence, or highly noxious conditions (e.g., janitor, mortician, soldier); work that is socially tainted involves interactions with stigmatized groups or is servile (e.g., correctional officer, chauffeur); and work that is morally tainted involves tasks that are perceived as sinful, deceptive, intrusive, or otherwise defying norms of civility (e.g., exotic dancer, collection agent)~\citep{ashforth_how_1999}.  

Although dirty work and dangerous work share common characteristics, social stigma is central to dirty work. Dirty work is fundamentally a social construct---its perceived value is context-dependent and shaped by social norms, values, and power dynamics~\citep{goffman_stigma_2009, ashforth_how_1999, simpson_introducing_2012}. This perspective means that social scientists studying dirty work often rely on qualitative methods (primarily using ethnography and in-depth interviews) to examine how society labels and stigmatizes certain forms of labor~\citep{ashforth_contextualizing_2014, e_ashforth_normalizing_2007}. This perspective also lends itself to cross-cultural and historical comparative studies to understand the variability of “dirtiness” of occupations (e.g., difference in social stigma around nurses in North America~\citep{mcmurray_embracing_2012} and Bangladesh~\citep{hadley_why_2007}, and changes over time with the stigma associated with the tattoo industry~\citep{Adams01032012}).

The literature on dirty work is also closely tied to scholarship on occupational prestige, which is defined as the perceived social status associated with different occupations~\citep{treiman_standard_1976}. Occupational prestige is typically quantified using surveys that ask respondents to either rank a list of occupations based on their prestige or rate them using a Likert-type scale~\citep{nakao_updating_1994, smith_measuring_2014, freeland_structure_2018, krueger_gender_2024}. For example, funeral directors and butchers both handle physical taint (death), but one is regarded to have relatively high occupational prestige (funeral director) while the other is regarded to have relatively low occupational prestige (butcher) in the United States. Although dirty work can have high prestige (e.g., surgeon, lobbyist), all else equal, dirty work occupations throughout the world are likely to have less prestige than non-dirty work occupations~\citep{ashforth_how_1999}. Although occupational prestige is often conflated with the education and income level of workers in the occupation, empirical research shows modest correlation between these measures and suggests the use of dimension-specific measures (i.e., separately measuring prestige, education, and income of a given occupation)~\citep{darity_jr_position_2022, hughes_occupational_2024, hauser_socioeconomic_1997}.

\subsubsection{Considerations for robotics} 

First, for many of us, social science research challenges our assumptions about dirty work. Social stigma is a helpful way to define dirty beyond just physical taint. It also adds complexity, because social stigma for different tasks can vary across time and culture.

Second, there may be opportunity to look for hidden dirty work that isn't immediately apparent to us. The literature on dirty work identifies a societal mechanism of moral outsourcing: the public benefits from those who perform dirty work, while the associated taint is externalized onto the workers doing dirty work. As a result, people who do not perform that work remain psychologically distant from the dirty work being performed, allowing them to remove themselves from the consequences of actions they implicitly condone or benefit from~\citep{hughes_good_1962}. Dirty work often remains hidden from public view, taking place overnight or spatially separated from the public (e.g., cleaners, slaughterhouse workers, prison guards). The robotics community may benefit from actively searching for this type of “dirty” labor that is often hidden from public view.

At the same time, it helps to recognize that it is sometimes automation itself that creates hidden work. In robotics, driverless vehicles rely on behind-the-scene teams of human technicians to remotely control the cars when they struggle to pilot themselves, but few outside of these companies were aware of the presence of human assistants when these driverless vehicles were deployed~\citep{metz_how_2024}. In the digital economy, companies employ human labor for content moderation, data labeling, and a host of other activities that are integral to the smooth operation of the business but remain invisible to customers~\citep{gray_ghost_2019}. This type of “ghost work” is pervasive in AI-related labor~\citep{williams_exploited_2022, miceli_data-production_2022}, continuing to fuel a moral outsourcing and lack of societal accountability for the human toll on workers doing dirty work. The robotics community should be mindful of its role---past and present---in creating this type of hidden labor, as well as the societal inequities that allow it~\citep{riek_future_2025}.

Third, there is a large body of qualitative research indicating that workers performing dirty work often derive meaning from their work and do not necessarily suffer from low self-esteem about their occupation~\citep{ashforth_how_1999}. Studies of sanitation workers~\citep{nagle_picking_2013} and truck drivers~\citep{karen_levy_data_2022} demonstrate that workers doing dirty work often take great pride in their labor, despite acute awareness of the stigma that attends to their profession. An anthropological study of Sri Lankan garment workers producing for export illustrates how work can be considered “dirty” due to social taint. In this case, societal concern about women's newfound work roles led to a stereotyping of garment workers as immoral. This study shows how, in spite of profound social stigma, women find meaning and community in their work~\citep{lynch_juki_2006}. These qualitative studies illustrate that the social stigma or physical threat associated with these jobs creates a sense of connection and camaraderie among the workers, fostering a positive association to the work and the communities that are built through this work.

Scholars have also demonstrated that some people take on dirty work because of the moral significance of providing care to others~\citep{simpson_introducing_2012}. These studies demonstrate that in order to prioritize automating tasks that are actually undesirable, it is important to account for and better understand how dirty work is experienced by workers performing these tainted tasks. 

\subsection{Dull Work}

\subsubsection{Social science review} “Dull work” is often considered synonymous with “drudgery,” a term that refers to \textbf{work that is repetitive and lacking in autonomy}. Political theorists have defined drudgery as uninteresting work, and “labor that does not involve self-realization, the exercise, development, and manifestation of one’s powers and capacities”~\citep{kandiyali_importance_2020}. Khandelwal notes that the term “drudgery” has been used to describe categories of work since at least the 1500s, and refers to “dull, irksome, and fatiguing work; uninspiring or menial labor”~\citep{khandelwal_what_2024}. Khandelwal also notes that Karl Marx, in his analysis of capitalism and industrial workplace processes, described work as “drudgery” that is repetitive, exhausting, and leads to alienation from work processes and products.  

Social scientists studying dull work have primarily focused their efforts on documenting workers' experiences and studying workers’ perspectives. As with the concepts of “dirty” and “dangerous,” outsiders may categorize as “dull” work activities that bring value and meaning to workers. In categorizing work, outsiders can miss much of what a job means to people doing the work. These outsider categories can serve as a form of judgment that flattens out and remove “layers of significance” that, from a worker's perspective, pertain to that work. Rose~\citep{rose_mind_2014}, cautions that our “occupational vocabularies” can be reductive and lead us to under-appreciate or fail to notice “many aspects of workplace intelligent behavior.” He also notes that “Many of our depictions of physical and service work… tend toward the one-dimensional.” For example, if, employing binary thinking, an outsider sees work either as “ennobling” or “dehumanizing,” the analysis may miss the meanings of work for people. Similarly, the outsider may erroneously assume that work is “mindless” if it is routinized. Rose describes how an outsider in an electrician’s workshop could mistake moments of quiet, routine concentration for “dull”~\citep{rose_mind_2014}. Just as Rose argues that we need to “reimagine and unsettle our prevailing vocabulary of work,” we encourage  the robotics community to do the same in order to understand what we miss when not accounting for the experiences and perspectives of workers.

\subsubsection{Considerations for robotics} 
\label{sec:sosci:consider-robotics}
First, understanding what constitutes “dull” work requires information gathered from close field observation and engagement with workers. For example, Khandelwal’s~\citep{khandelwal_what_2024} ethnographic study of the use of wood-burning stoves among women in India demonstrates the importance of understanding how people experience activities, tasks, and jobs, and taking note of what they value and why. Despite known health risks related to using wood stoves, women in the study use these stoves as a strategy to maximize flexibility and autonomy and to exhibit control. While walking miles to get wood for burning may appear to the outsider as boring work, full of drudgery, for some women this work offers an essential time for socialization and mutual support. Similarly, scholars of craft work have paid close attention to craft workers, showing that routine work, where one can focus on the work process, creates conditions for developing skill and competence~\citep{sennett_craftsman_2009}. These examples highlight that understanding the worker's experience reveals the  value and meaning of work. 

Second, the social science literature challenges the notion  that routine work must be eradicated. This presumption is based on a narrow understanding of how people work, think, learn, and make meaning. In an off-quoted 1951 piece about “dull work,” Hoffer, a philosopher and longshoreman, wrote: “There seems to be a general assumption that brilliant people cannot stand routine; that they need a varied, exciting life in order to do their best. It is also assumed that dull people are particularly suited for dull work....  Actually, there is no evidence that people who achieve much crave for, let alone live, eventful lives. The opposite is nearer the truth.... Einstein worked out his theory of relativity while serving as a clerk in a [S]wiss patent office”~\citep{hoffer_our_2008}.

This scholarship foregrounds an opportunity for robotics to move away from the presumption that repetitive work has only negative connotations. A key step is to examine how people experience their work and explore what opportunities a job offers to demonstrate autonomy, mastery, and purpose. Asking such questions may help us to avoid automating away desirable work, focusing instead on work that is truly experienced as undesirable ~\citep{pink_everyday_2022}.

\section{Framework for DDD}
\label{sec:framework}

Drawing on our review of the social science literature in Section~\ref{sec:socsci}, we propose a framework that provides definitions and guidance to examine whether work falls into the category of DDD. The framework is summarized in Figure 2 below. The Appendix includes a worksheet that guides users through this DDD framework.  

In this framework, work is understood in the context of both the \textbf{physical environment}, where physical tasks are performed, and the \textbf{social environment}, where workers interact with other people. For each term---dull, dirty, and dangerous---the framework gathers key pieces of information to reflect on what physical and/or social aspects of the work are, in fact, DDD:

\begin{figure}
  \includegraphics[width=\columnwidth, alt={The figure is a horizontal diagram that illustrates that examining jobs/tasks requires consideration of  physical and social environments of the jobs/tasks as well as questions about each job/task. The diagram consists of four overlapping rounded boxes arranged from left to right, each labeled with a concept and a guiding question: (1) Dangerous (lavender): “What information do we have about injuries/hazardous exposures from doing this job/task?” (2) Dirty (lavender): “What information do we have about the physical, social, or moral taint of this job/task?” (3) Dull (lavender): “What information do we have about routine activities/autonomy in this job/task?” (4) Workers (blue): “What information do we have about how workers themselves make sense of this job/task? Gather both positive and negative perspectives.” These rounded boxes lie on top of a grey background broken into two sections that are labeled “Physical environment" (top rectangle) and “Social environment” (bottom rectangle). The background and the rectangles are visually connected to show progression from physical to social considerations. Outside the boxes and grey background, two labels indicate dimensions of awareness. The two-way vertical axis is labeled: “Awareness of context dependence” (spanning the categories “Physical environment” and “Social environment.”) The two-way horizontal axis is labeled “Awareness of multiple perspectives” (spanning across the four rounded boxes).}]{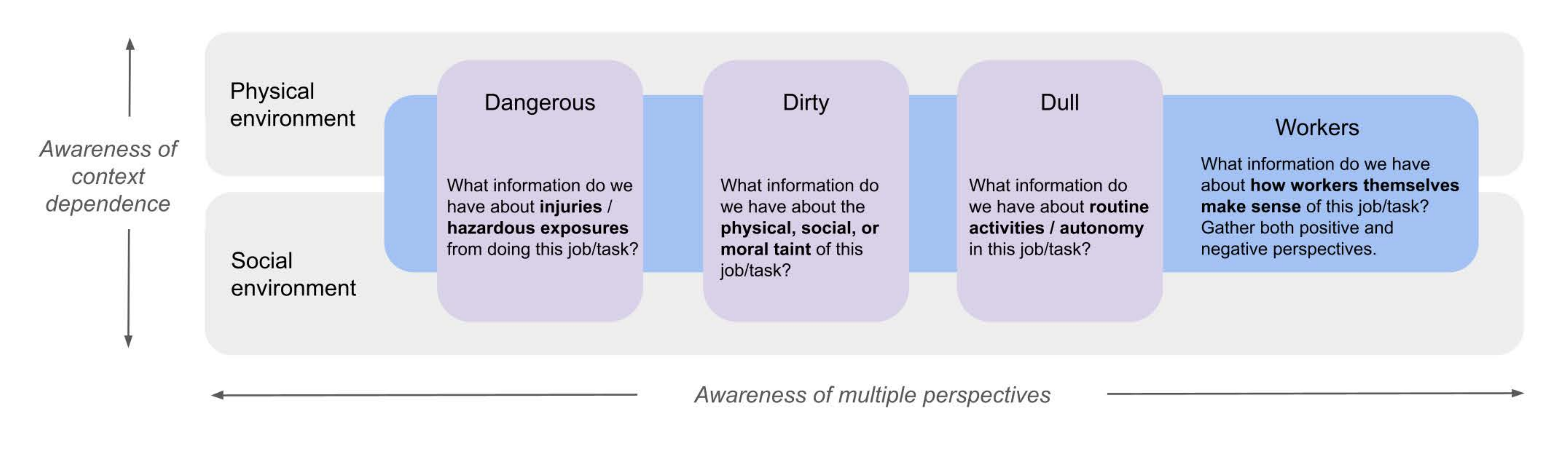}
  \caption{A DDD Framework for Examining Jobs/Tasks}
\end{figure}

\begin{itemize}
    \item \textbf{Dangerous}: What information do we have about injuries and hazardous exposures from doing this job?
        \begin{itemize}
            \item Examples: physical injury rates (physical) or exposure to harassment at work (social)
        \end{itemize}
    \item \textbf{Dirty}: What information do we have about the stigma associated with this job?
    \begin{itemize}
            \item Examples: handling of garbage, death, effluent (physical) or interacting with stigmatized groups (social)
        \end{itemize}
    \item \textbf{Dull}: What information do we have about routine activities and autonomy in this job? 
    \begin{itemize}
            \item Examples: experience working with repetitive motions (physical) or with little opportunity to interact with others (social)
        \end{itemize}
    
\end{itemize}

The framework encourages the collection of a range of quantitative and qualitative information, and consideration of potential sources of bias in the information. Ideally, someone looking to automate a task would gather information about the DDD nature of jobs under consideration, including how \textbf{workers themselves} make sense of that work. Here, both positive and negative perspectives from workers should be considered in order to paint a holistic picture of how that work is carried out in practice. The worker perspective is central to our framework and cuts across all three dull, dirty, and dangerous concepts. 

In addition to answering the questions, the framework attends to the following:

\begin{itemize}
    \item \textbf{Awareness of context dependence}: As noted in Section~\ref{sec:socsci}, the physical and social environment surrounding a job plays an important role in the extent to which that job is dull, dirty, or dangerous work. As a result, this framework encourages close attention to the broader context of that occupation and industry. This framework also encourages examining the positive dimensions of the physical and social environment of work, in ways that are not captured in the more narrow, negative, conventional use of DDD. 
    \item \textbf{Awareness of multiple perspectives}: As noted in  Section~\ref{sec:socsci}, social science literature highlights the importance of gathering a range of perspectives in order to have a more nuanced understanding of whether a job is rightly considered DDD. Some data may come from surveys or administrative records (for example, occupational injury rates for “dangerous” work), while other data may come from in-depth interviews, diary entries, or observations/ethnographies of workers doing the jobs (for example, qualitative data about the pride felt by workers doing socially stigmatized “dirty” jobs). Here, it is important to draw on a range of quantitative and qualitative data, and consider potential sources of bias.
\end{itemize}

\subsection{Applying the Framework to the Waste and Recycling Industry}
\label{sec:framework:example}

In this section, we briefly demonstrate how our framework can be applied to clarify what aspects of jobs/tasks may be considered dull, dirty, and/or dangerous. We show this by way of an example from the waste and recycling industry. The world currently generates over 2 billion tons of waste annually, and this figure is expected to rise to nearly 4 billion tons by 2050~\citep{silpa_kaza_more_2021}. Waste management is important for planning a sustainable and healthy future for all. 

\noindent \textbf{Dangerous}: \textit{What information do we have about injuries/hazardous exposures from doing this job?}

Globally, workers in the waste management and recycling industry face significant health hazards. Workers are exposed to toxic chemicals and biological agents and endure physical injuries such as falls, vehicle and object impact, cuts and lacerations, back injuries, and muscle strains~\citep{afanuh2024preventing, silpa_kaza_more_2021}. For example, in the United States, the occupational injury and illness incidence rates for refuse and recycling material collectors is 1.4 times higher than other transportation and material moving occupations and 3.6 times higher than the national average~\citep{united_states_bureau_of_labor_statistics_occupational_2024}.\footnote{For data gathering, we used the ``Refuse and Recyclable Materials Collector” occupation under the Standard Occupational Classification (SOC) code 53-7081.} Data from the International Labor Organization (ILO) and the World Bank suggest that injury rates in the waste and recycling industry are likely to be significantly higher in the Global South, where over 75\% of workers are estimated to be employed informally~\citep{silpa_kaza_more_2021}.\footnote{Official statistics on worker injury rates in the waste and recycling industry at the global scale are limited and likely to be underreported, as the majority of workers are employed informally.} Workers employed informally are often without basic social protections and employment benefits. In countries where occupational safety and health regulations are poorly enforced, workers are even more vulnerable to injuries and hazardous exposure~\citep{valentina_stoevska_beyond_2024}.

\noindent \textbf{Dirty}: \textit{What information do we have about the physical, social, or moral taint of this job?} 

Globally, waste collection is often seen as a low-status job and usually occupied by more marginal members of society. Data for the United States from the General Social Survey shows that people view refuse and recyclable collection workers as having relatively low occupational prestige, with a score of 34.22 on a scale of 0 (bottom) to 100 (top). This score ranks refuse and recyclable workers above embalmers (25.41) and below personal care aides (46.5) in terms of how the public views the prestige of these occupations~\citep{davern_2024}. The majority of solid waste in the world is collected in the Global South, and in many of these countries waste collectors are often migrants from rural areas and members of social minority groups. In India, waste collectors are more likely to be from the lowest caste (Dalits) and in many Muslim-majority countries, a higher share of non-Muslims take on the task of waste collection~\citep{anna_barford_call_2021}. Even within waste collection, higher status roles that collect more valuable materials (e.g., metals) tend to be occupied by men, whereas lower status roles that collect lower value materials (e.g., paper, glass) are typically occupied by women~\citep{noauthor_putting_2024, valentina_stoevska_beyond_2024}. 

How do workers make sense of the taint associated with the job? In an ethnography of New York City sanitation workers, Nagle~\citep{nagle_picking_2013} refers to an “invisibility syndrome”: sanitation workers know that they provide an essential service, but feel invisible to people for whom they provide that service. This invisibility is also noted by sanitation workers in Europe~\citep{hamilton2019lower} and across the Global South~\citep{maneen2025psychosocial}. For example, a study conducted in Bogotá, Colombia and Durban, South Africa finds that 80\% of waste pickers report repeated public harassment, and 97\% identify social exclusion as a significant issue in their daily work~\citep{noauthor_putting_2024}. Consequently, these workers often experience low levels of self-esteem. However, some sociological research suggests that workers in the refuse and recycling industry cope with this social stigma by reframing their work, choosing to take pride in the social accomplishments of protecting the environment and providing an essential service to the public~\citep{kreiner2022stigmatized, blustein2023understanding}. 

\noindent \textbf{Dull}: \textit{What information do we have about routine activities / autonomy in this job?}

Data from the US Department of Labor indicates that refuse and recycling collection workers often perform repetitive tasks, with 57\% of workers responding that they continually spend time making repetitive motions on the job. Workers also report limited autonomy on the job, with only 39\% of workers responding affirmatively to having freedom to determine tasks, priorities, or goals~\citep{us_department_of_labor_onet_nodate}.

How do workers make sense of the routine activities and autonomy in this job? Qualitative studies of refuse and recycling collection workers suggest that workers often find their routes and schedules to be repetitive. However, workers also find enjoyment in their job. Specifically, workers cite the day-to-day interaction with people on their routes~\citep{deery2019can} and with their coworkers~\citep{nagle_picking_2013}, which are manifested in their extensive insider vocabulary and work hacks, and in long-standing employee mutual aid groups that support each other in times of need. Nagle found camaraderie among New York City sanitation workers built through daily enactment of "house rules," or insider forms of expertise that counter formal managerial regulations~\citep{nagle_picking_2013}. Workers cite task variety as one of the most captivating aspects of the job~\citep{perry2017}. The range of tasks vary considerably, from inspecting their vehicle and equipment, driving their truck, coordinating with crew members, lifting bins and bags, detecting incorrect sorting of waste, to unloading at the end destination.  

After going through our DDD framework, we have a range of data sources that highlight the aspects of the job that are dull, dirty, and/or dangerous. In the waste and recycling industry example, we found a range of evidence about the extent to which the work of refuse and recycling material collectors is dangerous, the stigma associated with the job, and ways in which the work is perceived to be dull---both by workers themselves and by others in society.

The collected data indicate that sanitation work is dangerous and dirty relative to other occupations, and some parts of the job are dull to workers themselves. However, the data also shows that there are aspects of the job that workers value: task variety, social interaction with coworkers and the public, and a sense of autonomy. 

This matters, because introducing some types of robotic solutions will inherently reduce task variety for incumbent workers. For instance, consider the adoption of automated systems recommended by the National Institute for Occupational Safety and Health (NIOSH). NIOSH recommends the adoption of automated side loader trucks aided by sensors and cameras to pick up and dump containers (to prevent back injuries and muscle strains), as well as the installation of proximity detection and collision avoidance systems (to protect workers from vehicle accidents)~\citep{afanuh2024preventing}. Implementing these recommendations would address safety concerns, but would also lead to reduction of task variety. For example, side-loading trucks require only one worker (the driver) and eliminate the need for a worker to step off trucks to handle bins: now the sole worker operates a joystick in the cab to direct the hydraulic arm. Another potential downstream consequence of implementing these recommendations would be a reduction in social interaction (no more need for the verbal and non-verbal communication that has contributed to workplace camaraderie in the industry) and autonomy (managerial surveillance via sensors and cameras can create worker concern about always being watched, as noted for the introduction of GPS technology in the industry~\citep{nagle_how_2011}). Our framework reveals these tradeoffs, highlighting an opportunity to consider what tasks to prioritize in designing automation solutions. It challenges us to ask: How might we both reduce the incidents of injury while also retaining or creating new forms of autonomy and social interaction? Our framework provides an accessible and actionable set of questions for roboticists to examine work under the lens of DDD.

\section{Limitations \& Future Work}

In this paper, we conducted an empirical review of robotics papers to examine the use of DDD. Our review was limited to peer-reviewed papers and did not examine the use of DDD more broadly in industry or government.  

We proposed an accessible and actionable DDD framework. However, research has shown that the widespread use of these kinds of resources (e.g., toolkits, frameworks) is limited by the amount of effort required from users to adapt it to their context~\cite{deng_exploring_2022}. To avoid the burden of requiring users of our framework to collect new primary data, our framework encourages users to take advantage of the extensive and valuable worker data already collected by social scientists. A future direction of this work could be primary data collection in relevant areas where existing data is lacking, such as disaggregated data on gender (as discussed in Section~\ref{sec:sosci:consider-robotics}). 

Future work can focus on how to simplify gathering information about dull, dirty, and dangerous aspects of tasks and jobs. To this end, our team is developing a database that links various data sources under our DDD framework to help streamline the information-gathering process. To inform database design, we plan to conduct a user study with roboticists to examine how they apply the framework.

\section{Conclusion}

In this paper, we reviewed publications in robotics between 1980 and 2024 and have shown that while the term “dull, dirty, dangerous” (DDD) is widely used to motivate research, very few papers define these terms or provide provenance citations. Only 8.7\% of reviewed papers provided a concrete example of a task or job that the authors categorized as DDD, and a mere 2.7\% of papers provided a definition of the specific terms.

 The lack of definition around DDD can create an illusion of consensus among roboticists about what tasks and occupations are DDD. In other words, roboticists may be basing DDD determination on personal beliefs and ideas about occupations, rather than on an informed understanding of the types of tasks and occupations where robotics can add value to society. Our social science literature review establishes definitions and measures that add robustness to the terms as applied to robotics. Finally, our DDD framework offers an actionable way for the robotics community to think about how tasks and jobs will be impacted by technology. Our framework encourages us to engage with multiple sources of information---most importantly, workers’ perspectives---to consider how our automation choices can be guided by real-world data.
 
Because much of robotics development and deployment is task-specific (versus general purpose), we believe it is important for roboticists to consider the job context that technology will impact. As a community, we have an opportunity to identify and focus efforts on making technical advancements that can solve meaningful problems and create more positive experiences in the world.

\section*{Acknowledgements}
We would like to thank members of the Robotics, Ethics \& Society team, Jenny Barry, Maximillian Clee, Jessica Hodgins, Marc Raibert, and the conference referees for their helpful comments and feedback on the project.

\bibliographystyle{unsrtnat}
\bibliography{references}  






\clearpage

\section{Supplementary Materials}

\renewcommand{\thefigure}{S\arabic{figure}}
\renewcommand{\thetable}{S\arabic{table}}
\setcounter{figure}{0} 

\begin{table}[h]
  \centering
  \caption{Definitions of Dull, Dirty, and Dangerous found in the paper sample collected in our dataset}
  \addtocounter{table}{-1} 
  \small
  \label{tab:commands}
  \begin{tabular}{ccp{8cm}}
    \toprule
    \textbf{Author(s)} & \textbf{Year} & \textbf{Definitions} \\
    \midrule
        Clark & 2000 & “Dull means long-endurance missions that can last for several hours or even days—too long for a pilot to physically endure. Missions such as extended patrols over no-fly zones would fall into this category, or missions that require long-endurance flights to get to the target area and back would also qualify. Dirty missions are those where the threat of biological or chemical contamination is too high to risk sending in a manned aircraft. Dangerous missions for manned aircraft are numerous and growing; and though most combat missions are dangerous, some are riskier than others.” \\ \hline
        De Sousa and Pereira & 2002 & “‘Dirty’ refers to reconnoitering areas that may be contaminated, ‘dull’ applies to surveillance or sentry duty, while ‘dangerous’ is related to obvious threats, such as those posed by the suppression of enemy air defenses.” \\ \hline
        Austin & 2003 & “Dull Roles - Military and civilian applications such as extended surveillance can be a dulling experience for aircrew, with many hours spent on watch without relief, and can lead to a loss of concentration and therefore loss of mission efficiency. Dirty Roles - Again, applicable to both civilian and military applications, monitoring the environment for nuclear or chemical contamination puts aircrew unnecessarily at risk. Dangerous Roles - For military roles, where reconnaissance of heavily defended areas is necessary, the attrition rate of a manned aircraft, with its higher probability of detection, will exceed that of a RR. Powerline Inspection is an example in the civilian field where experience sadly has shown that aircrew are at significant danger.” \\ \hline
        Carafano et al. & 2008 & “Dull assignments are those that require routine functions such as monitoring a bridge crossing site. Dirty jobs are performed in harsh environmental conditions, such as searching contaminated areas. Dangerous missions involve tasks in which humans could suffer physical harm, such as disarming an improvised explosive device (IED).” \\ \hline
        Salazar et al. & 2008 & “Examples include orbiting endlessly over a point for communications relay or jamming (dull), collecting air samples to measure pollution or CW/BW toxins (dirty), and ﬂying reconnaissance over hostile air defenses (dangerous).” \\ \hline
        Deputy & 2009 & “The term dull within the three ‘D’s refers to complacency, or shortcomings in human attention and endurance.” \\ \hline
    \end{tabular}
\end{table}

\begin{table}
  \centering
  \caption{Continued}
  \addtocounter{table}{-1} 
  \small
  \label{tab:commands}
  \begin{tabular}{ccp{8cm}}
    \toprule
    \textbf{Author(s)} & \textbf{Year} & \textbf{Definitions} \\
    \midrule
    Hayashi et al. & 2011 & “Dull task: Endless Guidance: Guidance is one common communicative task in urban environments at large shopping malls and event sites. Providing guidance just once is simple and not so dull; however, it quickly becomes monotonous if the guidance task is endless, e.g., providing continuous guidance for more than a few hours. This situation is actually common. Such tasks are currently assigned to human guides in such situations as special events at amusement parks. Thus, we chose an endless guidance task as an example of a dull task. Dirty Task: Picking a Key Out of Garbage: In a city, cleaning toilets and garbage collection are typically dirty tasks, some of which are already being done by robots. For them, people naturally prefer robots.” \\ \hline
    Ozalp and Sahingoz & 2013 & “Dangerous environment term is used for operationally risky areas for both aircrafts and pilots. Dirty defines especially the environment which may be contaminated by chemical, biological, or nuclear particles. Finally, dull is used for tasks, which require long lasting operations, which are not desired to use manned planes.” \\ \hline
    Pascarella et al. & 2013 & “Dull operations are too monotonous or require excessive endurance for human occupants. Dirty operations are hazardous and could pose a health risk to a human crew. Dangerous operations could result in the loss of life for the on-board pilot.” \\ \hline
    Cakir & 2015 & “The term of dangerous environment is used for operationally risky areas for both aircrafts and pilots. Dirty deﬁnes especially the environment which may be contaminated by chemical, biological, or nuclear particles. Finally, dull is used for tasks, which require long lasting operations, which are not desired to use manned planes.” \\ \hline
    Dave and Patil & 2015 & “Dull Applications, Dirty applications are those not best suited for a human operator due to the circumstances in which they are to be undertaken.” \\ \hline
    De Boisboissel & 2015 & “Dangerous: in the case of dangerous threats, it is of the utmost importance to automate certain actions, enabling us to satisfactorily handle several targets, handle a target before it deals with us or before it handles friendly elements. Quick reaction times must prevail. Dull: for extremely repetitive tasks, humans rely on machines with a technology capable of detecting threats or alarms with greater accuracy and steadfastness than any human could ever reach whilst obviating physical tiredness and mental fatigue. In the case of surveillance missions in high risk areas where open access is strictly forbidden (nuclear missile warehouses, No Man’s Land zones, etc.) quick reaction times are also essential. Dirty: it is very strenuous for military units to operate in polluted and contaminated environments as CBRNs4.” \\ \hline
    \end{tabular}
\end{table}

\begin{table}
  \centering
  \caption{Continued}
  \addtocounter{table}{-1} 
  \small
  \label{tab:commands}
  \begin{tabular}{ccp{8cm}}
    \toprule
    \textbf{Author(s)} & \textbf{Year} & \textbf{Definitions} \\
    \midrule
    Hirukawa & 2015 & “…it is dangerous for a human to carry a heavy object.... The working space is dangerous for humans under some conditions including high temperature, a chemical material, a radiation and geographical features.” \\ \hline
    Bartsch & 2015 & “Any current activity in which airborne assets are deployed in a ‘dull’ (long endurance, high altitude, fatigue inducing), ‘dirty’ (volcanic plumes, chemical spills) or ‘dangerous’ (high risk, low altitude such as firefighting) environment may potentially be conducted in a safer, less expensive, and more efficient manner with unmanned aircraft.” \\ \hline
    Kaplan et al. & 2016 & “Dull, long-duration missions” \\ \hline 
    Fahey and Luqi & 2016 & “Dull missions tax a human’s concentration, are subject to fatigue forces, and require prolonged vigilance. This mission area is better served by an entity that does not get tired, frustrated, or bored, that will make ‘good’ decisions twenty minutes or twenty hours into a sortie. Dirty missions involve exposure to unhealthy environmental conditions, such as fumes, toxic substances, infectious biological materials or radiation. Entities that are not sensitive to such conditions can perform such missions with fewer constraints on exposure time.” \\ \hline
    Okcu & 2016 & “The term dull is used for repetitive and long tasks, which personnel suffer from. Secondly, the term dirty is used for the operation areas, in which personnel cannot enter due to the hazards such as radiation, biological and chemicals agents etc. Finally, the term dangerous means that the operation has great risks for human life.” \\ \hline
    Cullen et al. & 2017 & “Dull: Tasks with low workload and low intensity such as surveillance over a fixed location. These tasks assume a degree of automation, but without continuous direct control, or sophisticated autonomy, any high-priority opportunity which arises may be missed. Dirty: Operations in environments which are hostile to a manned aircraft or its crew. An example of this would be observation related to chemical, biological, radiological, or nuclear detection. Dangerous: Tasks where, for example, there is a high ground to air threat which does not merit the risk to aircrew or soldiers on the ground.” \\ \hline
    Guss et al. & 2017 & “The ‘dull’ aspect refers to repetitive missions. ‘Dirty’ refers to environments in which there are nuclear, biological, and chemical threats. The tasks deemed ‘dangerous’ are those in which there is a high risk to the aircraft and aircrew.” \\ \hline
    Luria et al. & 2020 & “social roles that users may not be interested in taking on themselves” \\ \hline
    Rath et al. & 2023 & “…grimy (Dirty) , Dull, or risky (Dangerous / Dicey)” \\ \hline
    \end{tabular}
\end{table}

\begin{table}
  \centering
  \caption{Continued}
  \addtocounter{table}{-1} 
  \small
  \label{tab:commands}
  \begin{tabular}{ccp{8cm}}
    \toprule
    \textbf{Author(s)} & \textbf{Year} & \textbf{Definitions} \\
    \midrule
    Toledano-García et al. & 2023 & “Dull: The task is repetitive or tedious; Dirty: The task or the environment where it is developed is unpleasant; Dangerous: The task or the environment represents a security risk.” \\ \hline
    Gómez-DelaHiz et al. & 2024 & “…triple D ‘Dull, Dirty, and Dangerous’ which tries to describe their potential and usefulness, since they can substitute other beings in ‘dull’ (long-lasting, slow or even unimportant activities), ‘dirty’ (access to volcanoes, agricultural tasks or handling of toxic materials) and ‘dangerous’ (assistance in natural disasters) activities.” \\ \hline
    Kutzke et al. & 2024 & “This is due in part to the benefit of allowing robots to take on the dirty, dull, and dangerous tasks that would otherwise place humans in direct harm or assist in situations in which it is not possible for the human to provide the level of persistent attention required, as in the case of long-term care facilities.” \\ \hline
    Mochurad et al. & 2024 & “Dangerous means that the life of a pilot can be at high risk during the flight. Dirty implies that the operational environment may be contaminated with chemical, biological, or radiation threats, making it unsafe for humans. Dull refers to long-duration, exhausting, and stressful operations that are undesirable for human performance.” \\ \hline
    \end{tabular}
\end{table}

\begin{figure*}[h]
    \centering
    \includegraphics[width=\linewidth, alt={Figure is a horizontal stacked bar chart showing the distribution of papers across academic fields and their rates of citing or defining the Dull, Dirty, and Dangerous framework. The vertical axis shows Academic Field with 13 categories. The horizontal axis shows Number of Papers ranging from 0 to 700. Three metrics are displayed: Total number of papers (blue), number of papers with citations for DDD (orange), and number of papers with DDD definitions (green), with percentages shown for each. The data are summarized in the following table. Design: 2 total papers (0.2 percent), 0 with citations (0.0 percent), 0 with definitions (0.0 percent). Engineering: 680 total papers (74.0 percent), 165 with citations (24.3 percent), 16 with definitions (2.4 percent). Ethics: 4 total papers (0.4 percent), 2 with citations (50.0 percent), 0 with definitions (0.0 percent). HCI: 13 total papers (1.4 percent), 1 with citations (7.7 percent), 0 with definitions (0.0 percent). Healthcare: 1 total paper (0.1 percent), 0 with citations (0.0 percent), 0 with definitions (0.0 percent). HRI: 45 total papers (4.9 percent), 17 with citations (37.8 percent), 0 with definitions (0.0 percent). Information Science: 1 total paper (0.1 percent), 0 with citations (0.0 percent), 0 with definitions (0.0 percent). Military: 133 total papers (14.5 percent), 39 with citations (29.3 percent), 8 with definitions (6.0 percent). Robotics: 360 total papers (39.2 percent), 78 with citations (21.7 percent), 9 with definitions (2.5 percent). Sensing: 13 total papers (1.4 percent), 3 with citations (23.1 percent), 0 with definitions (0.0 percent). Social Robotics: 2 total papers (0.2 percent), 1 with citations (50.0 percent), 1 with definitions (50.0 percent). Software: 1 total paper (0.1 percent), 1 with citations (100.0 percent), 0 with definitions (0.0 percent). STS: 2 total papers (0.2 percent), 1 with citations (50.0 percent), 0 with definitions (0.0 percent). Engineering and robotics dominate the corpus. Citation rates vary considerably by field, with HRI showing the highest citation rate at 37.8 percent. Definition rates remain low across all fields, with military showing the highest at 6.0 percent.}]{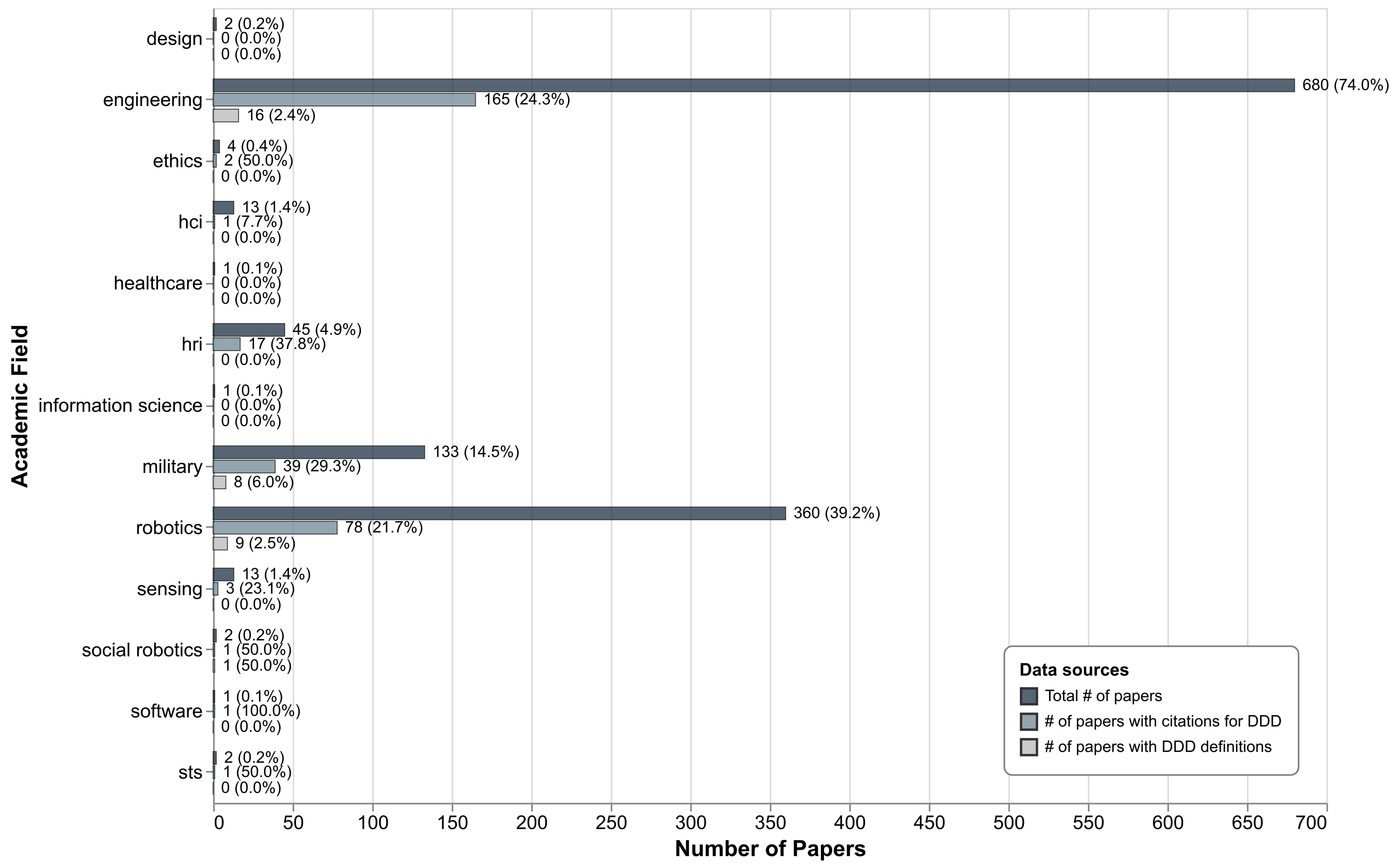}
    \caption{Absolute values and relative percentages of papers per field, papers offering citations supporting DDD concepts, and papers providing definitions of DDD concepts}
    \label{fig:citation_field}
\end{figure*}

\begin{figure*}
\centering
    \includegraphics[scale=0.3, alt={Figure is a PRISMA flow diagram showing the systematic literature review process for articles on Dull, Dirty, and Dangerous robotics across three stages: Identification, Screening, and Included. Identification stage: Articles identified based on queries (n=1418) from ACM DL (n=67), IEEE Xplore (n=522), and Dimensions (n=829). The queries used were: ACM DL searched All fields for dirty AND dull AND dangerous AND robot, filtered to Research Articles; IEEE Xplore searched Full Text and Metadata for dirty AND dull AND dangerous AND robot, filtered to Conferences and Journals; Dimensions searched for dirty dull dangerous within 2 words AND robot, filtered to Engineering or Information and Computing Sciences. Screening stage: Records removed before screening with duplicate records removed (n=8). Full text articles assessed for eligibility (n=1410). Included stage: Full text articles excluded (n=491) from ACM DL (n=23), IEEE Xplore (n=70), and Dimensions (n=398). Full text articles included (n=919) from ACM DL (n=43), IEEE Xplore (n=445), and Dimensions (n=431). The diagram shows that approximately 65 percent of assessed articles were included in the final corpus after removing duplicates and ineligible articles.}]{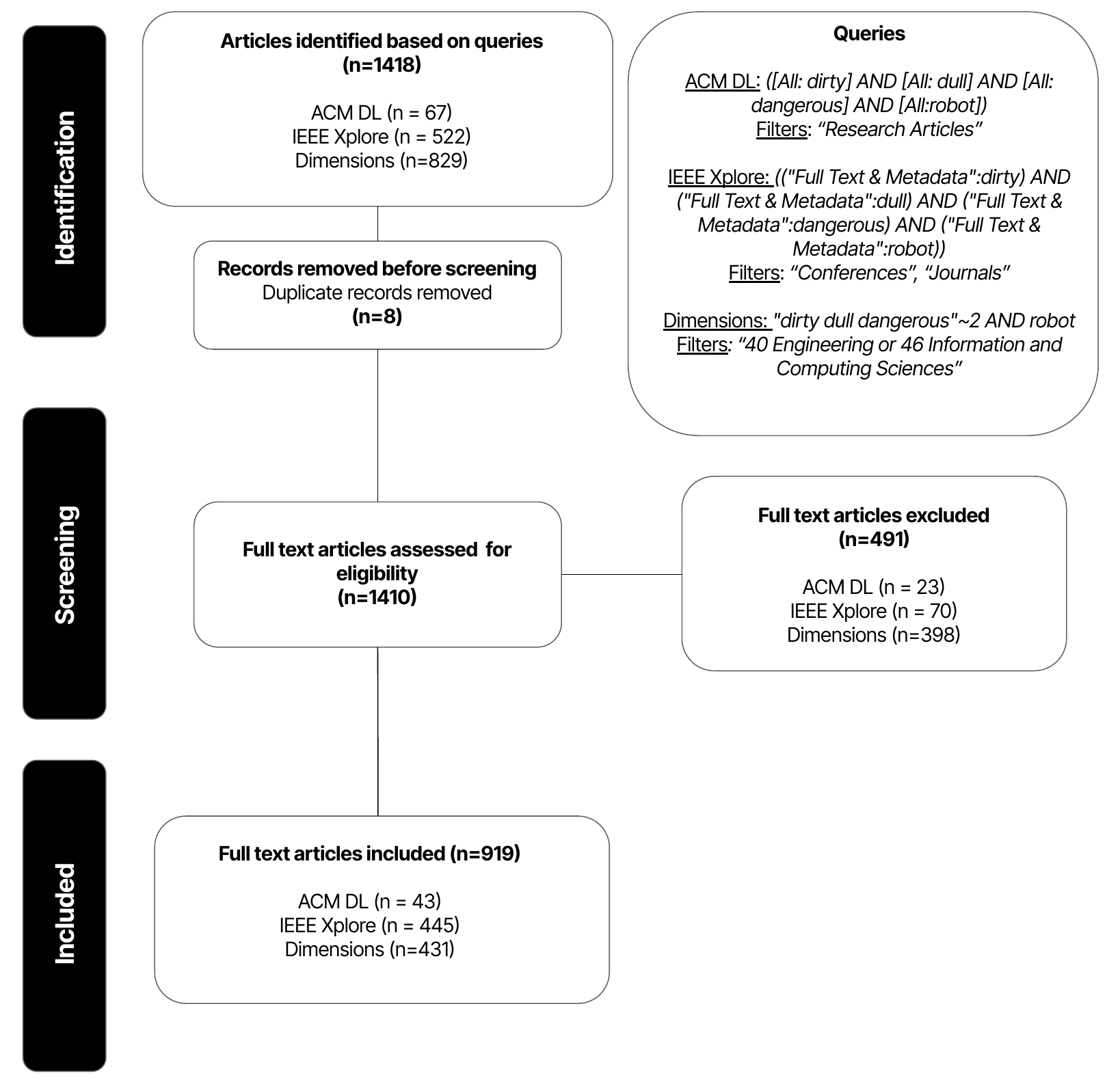}
    \caption{Prisma flowchart diagram showing the process of selecting items for eligibility into the dataset}
    \label{fig:prisma}
\end{figure*}

\begin{figure*}
    \begin{center}
    \includegraphics[scale=0.25, alt={Figure is a DDD Worksheet designed for users to explore the suitability of robotics applications and assess work using the Dull, Dirty, and Dangerous framework. The worksheet contains several sections. Instructions: Explains that users will gather information about each concept (dangerous, dirty, dull) and collect worker perspectives. Fill-out fields: Spaces for Task/Job and Industry of task/job. Definitions section: Dangerous work is occupations or tasks that result in injury or risk of harm. Dirty work is occupations or tasks that are physically, socially, or morally tainted. Dull work is occupations or tasks that are repetitive or lacking in autonomy. Conceptual framework diagram: Shows progression from "Physical Environment" to "Social Environment" with four numbered boxes: (1) "Dangerous" asking about injuries/hazardous exposures, (2) "Dirty" asking about physical, social, or moral taint, (3) "Dull" asking about routine activities/autonomy, (4) "Workers" asking about worker sense-making with both positive and negative perspectives. The diagram has two axes: "Awareness of context dependence" (vertical) and "Awareness of multiple perspectives" (horizontal). This is the same figure as Figure 2 in the main text. Question section: Four questions corresponding to the framework boxes, each with space for responses. Checklist of considerations: Multiple checkboxes organized under two main categories: Multiple perspectives (Did you search for existing administrative data, surveys, in-depth interviews, diary entries, observations/ethnographies?) and Context dependence (Did you consider the physical and social environment and broader industry context?). Additional prompts ask about including quantitative and qualitative information, considering sources of bias, and including positive and negative worker perspectives. Example references listed include: International Labour Organization database, US Department of Labor surveys, European Skills database, US O*NET Database, and US General Social Survey, plus journal articles and books from anthropology, sociology, and labor studies. The worksheet provides a structured methodology for comprehensive work assessment that moves beyond physical hazards to include social context and worker experiences.}]{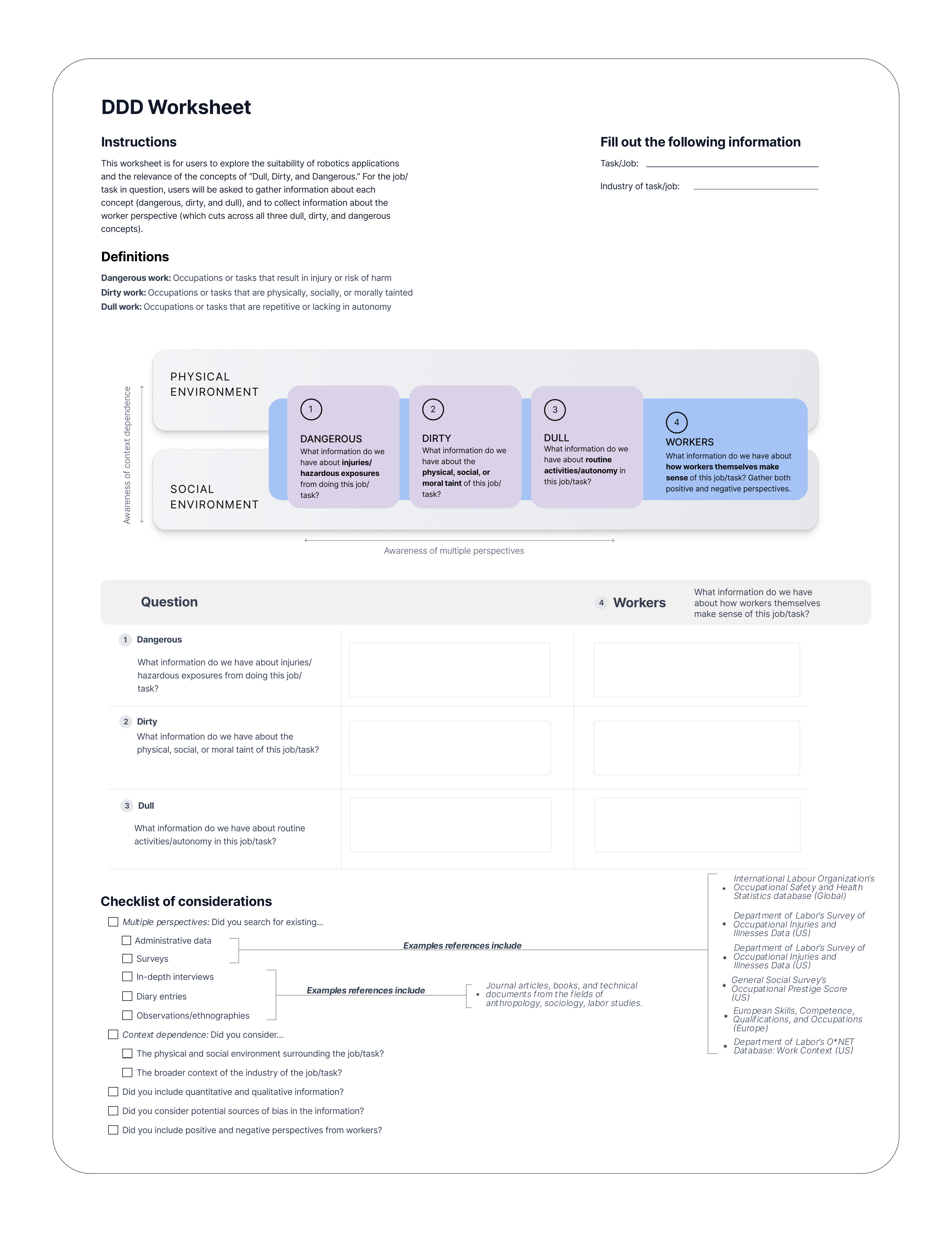}
    \caption{One-page worksheet to guide users through our DDD framework}
    \label{fig:worksheet}
    \end{center}
\end{figure*}

\end{document}